\documentclass[letterpaper]{article} 
\usepackage{aaai25}  
\nocopyright
\usepackage{times}  
\usepackage{helvet}  
\usepackage{courier}  
\usepackage[hyphens]{url}  
\usepackage{graphicx} 
\usepackage{natbib}  
\usepackage{caption} 
\usepackage{algorithm}
\usepackage{algorithmic}
\usepackage{subfigure}
\usepackage{times}
\usepackage{latexsym}
\usepackage{amsmath}
\usepackage{amssymb}
\usepackage{graphicx}
\usepackage{booktabs}
\usepackage{multicol}
\usepackage{multirow}
\usepackage{amsmath}
\usepackage{amsfonts} 
\usepackage{subfigure}
\usepackage{marvosym}

\usepackage{newfloat}
\usepackage{listings}
\DeclareCaptionStyle{ruled}{labelfont=normalfont,labelsep=colon,strut=off} 
\floatstyle{ruled}
\newfloat{listing}{tb}{lst}{}
\floatname{listing}{Listing}
\title{Adaptive Reinforcement Learning Planning: Harnessing Large \\ Language Models for Complex Information Extraction}
\author{
Zepeng Ding$^\diamondsuit$$^\heartsuit$, Ruiyang Ke$^\diamondsuit$$^\spadesuit$, Wenhao Huang$^\diamondsuit$$^\spadesuit$, \\ Guochao Jiang$^\diamondsuit$$^\heartsuit$, {\bf Yanda Li}$^\diamondsuit$$^\heartsuit$, {\bf Deqing Yang}$^\diamondsuit$$^\heartsuit$, {\bf Jiaqing Liang}$^\diamondsuit$$^\heartsuit$\textsuperscript{\Letter}
}
\affiliations{
$^\diamondsuit$Shanghai Key Laboratory of Data Science \\$^\heartsuit$School of Data Science, $^\spadesuit$School of Computer Science, Fudan University \\

\texttt{\{zpding22, ryke23, whhuang21, gcjiang22, ydli22\}@m.fudan.edu.cn} \\
         \texttt{\{yangdeqing, shawyh, liangjiaqing\}@fudan.edu.cn}\\
}

\usepackage{bibentry}

\begin{document}

\maketitle

\begin{abstract}
Existing research on large language models (LLMs) shows that they can solve information extraction tasks through multi-step planning. However, their extraction behavior on complex sentences and tasks is unstable, emerging issues such as false positives and missing elements.
We observe that decomposing complex extraction tasks and extracting them step by step can effectively improve LLMs' performance, and the extraction orders of entities significantly affect the final results of LLMs. This paper proposes a two-stage multi-step method for LLM-based information extraction and adopts the RL framework to execute the multi-step planning. We regard sequential extraction as a Markov decision process, build an LLM-based extraction environment, design a decision module to adaptively provide the optimal order for sequential entity extraction on different sentences, and utilize the DDQN algorithm to train the decision model. We also design the rewards and evaluation metrics suitable for the extraction results of LLMs. We conduct extensive experiments on multiple public datasets to demonstrate the effectiveness of our method in improving the information extraction capabilities of LLMs.
\end{abstract}

%

\section{Introduction}

Current large language models (LLMs) have excellent performance in many fields such as dialogue, reading comprehension, and text generation, greatly expanding the boundaries of traditional NLP tasks and thus demonstrating the capacity to support complex information extraction (IE) tasks.
Large models can understand extraction requirements through specific prompts and improve their performance on downstream tasks through in-context learning, multi-step planning \cite{wei2023zero}, collaborating with small model \cite{ding2024improving} and so on. These methods achieve good performance in general IE tasks.

\begin{figure}[!htb]
\begin{center}
\subfigure[]{
\includegraphics[width=0.95\linewidth]{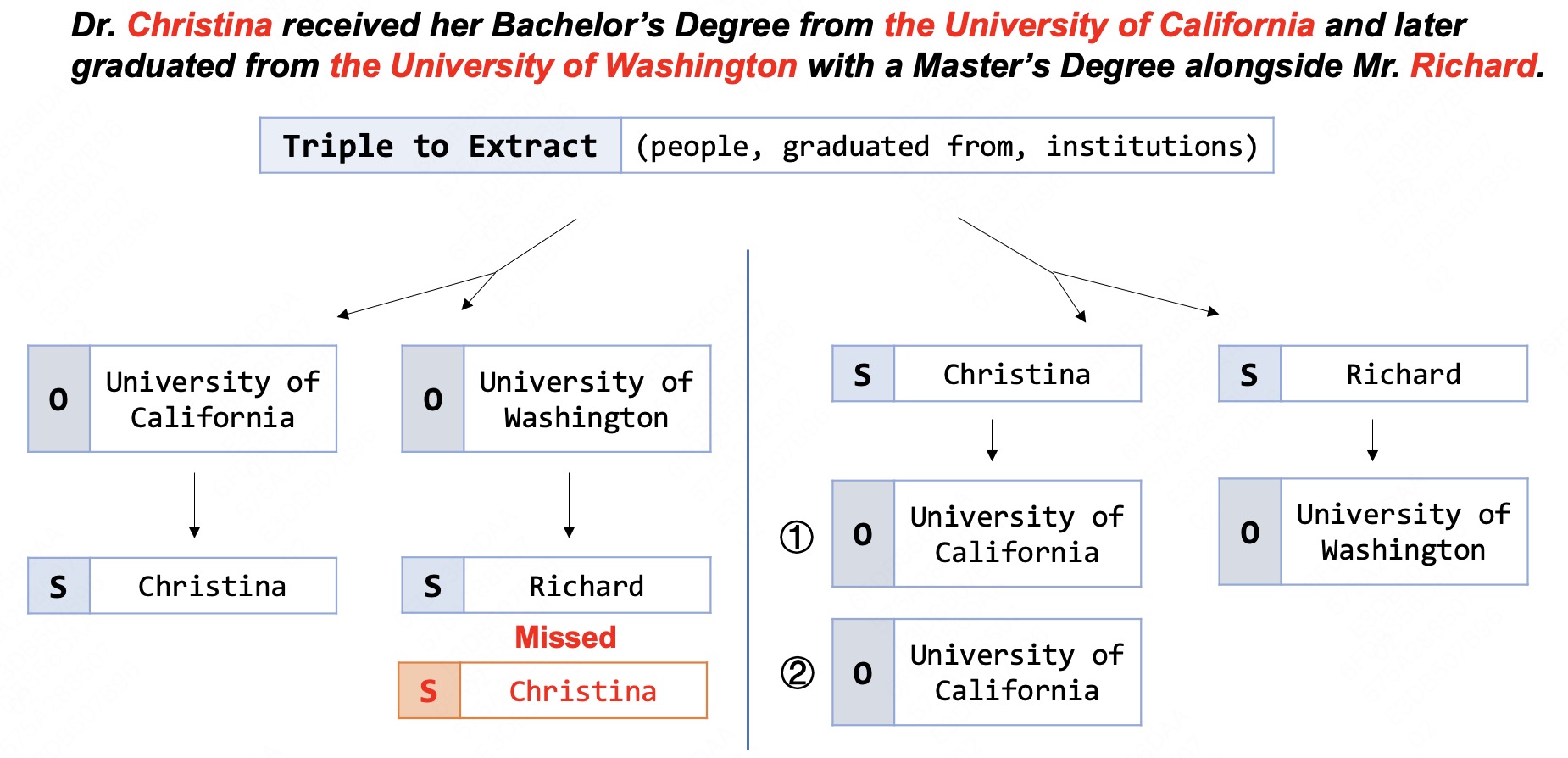}
\label{fig:moti_1}
} \\
\subfigure[]{
\includegraphics[width=0.95\linewidth]{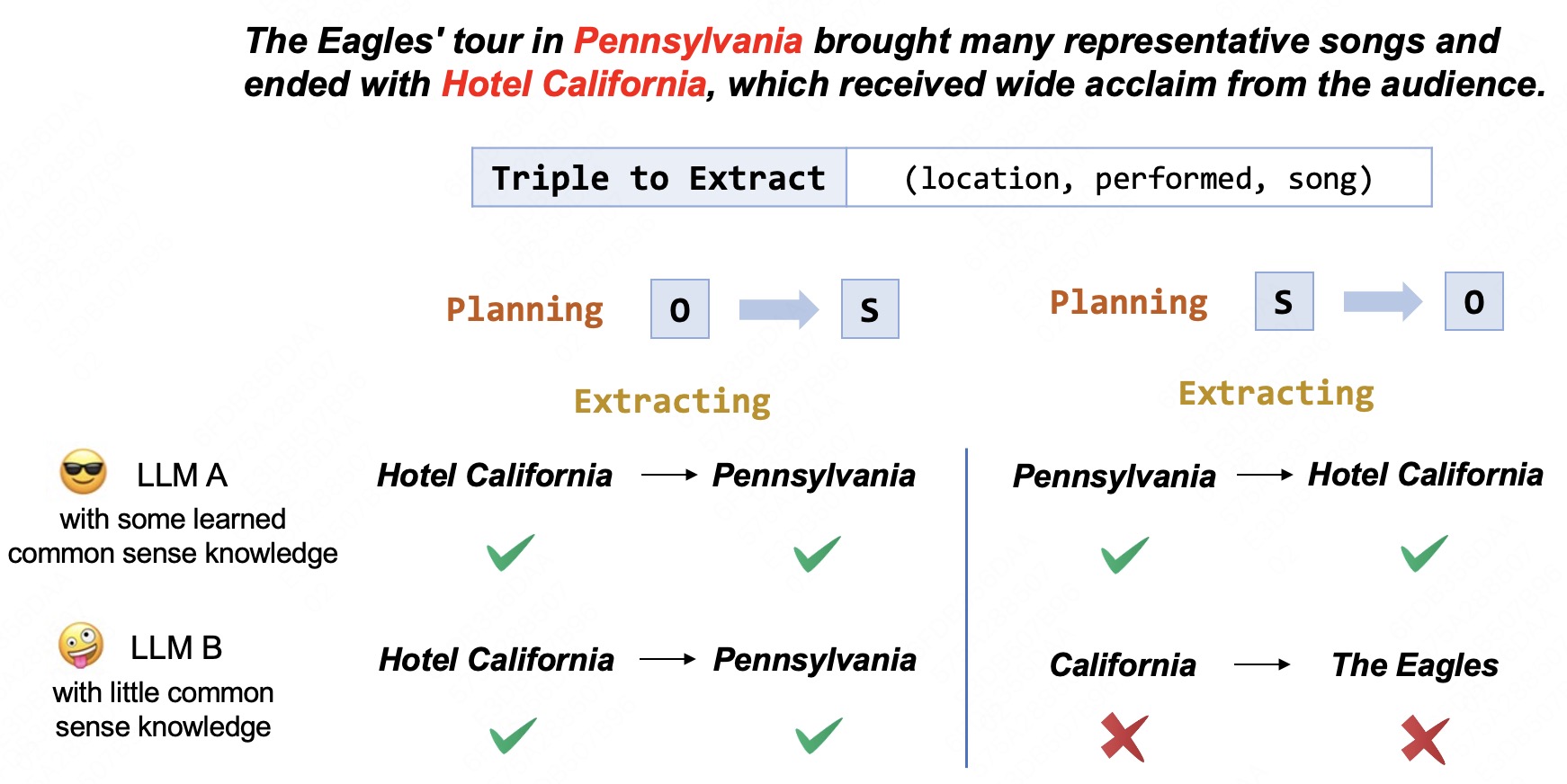}
\label{fig:moti_2}
}
\caption{(a) Extraction orders influence the outputs.  (b) Planning and extracting are entangled for each model.}
\label{fig:moti}
\end{center}
\end{figure}

However, when confronted with complex situations such as very long sentences or paragraphs, multiple related entities in the same sentence, and tokenization confusion caused by overlapping tokens, LLMs still engender issues such as false positives and missing elements \cite{ma2023large,wang2024tokenization}, resulting relatively low F1-score. Fine-tuning LLMs with task data is a common solution to enhance extraction capabilities, but it is time-consuming, has limited generality and its performance highly relies on the quality of the labeled data. Even though LLMs are powerful, they still require multiple steps to achieve acceptable extraction results for complex cases  \cite{wei2023zero}.

Previous studies on small language models (SLMs) show that task decomposition and sequential extraction could reduce errors and achieve better results \cite{xie-etal-2021-revisiting, lu-etal-2022-unified}. 
In our work, we divide the extraction process of LLMs into two distinct tasks: the classification of relations/events and the extraction of entities (arguments). We extract entities in multiple steps, \textbf{orderly} rather than concurrently.
Indeed, many studies on LLM-based agents are using large models to solve problems in an orderly step-by-step manner with interaction and feedback, such as ReAct \cite{yao2022react} and AutoCrawler \cite{huang2024autocrawler}. The effectiveness of these methods \textbf{heavily depends on the inherent planning ability of LLMs}.

Taking complex information extraction as an example, many studies prove that it requires particularly high planning capabilities, which poses a considerable challenge to the aforementioned agents. 
First, \textbf{the results of multi-step extraction are sensitive to the order of steps}. \citet{huang-etal-2023-adaptive} verifies that the extraction order affects the results of the BERT-based model, but the structure of the generative large model and its performance on language tasks differ from those of the traditional small model, so the extraction performance of LLMs deserves further study\footnote{In IE tasks, traditional models tend to extract excessive entities, but LLMs tend to get a low recall result \cite{ding2024improving}.}.
Previous works on LLMs use preset prompt templates and present a fixed order planning \cite{wei2023zero,ding2024improving}. However, we find that varying extraction orders significantly influences the output of LLMs, and the optimal order of entity extraction is not fixed for different sentences. 
As illustrated in Fig.~\ref{fig:moti_1}, different extraction orders lead to different results, and the optimal order is distinct from the situation in Fig.~\ref{fig:moti_2}. 
Second, the optimal order for models of different capabilities and types is different, making \textbf{planning and extraction execution entangled and difficult to decompose}. If the previous planning causes extraction errors in the intermediate results, the wrong results will also affect the execution of subsequent step(s).
As illustrated in Fig.~\ref{fig:moti_2}, LLM A can identify the song ``\textit{Hotel California}" based on its inherent common sense knowledge, so both plans can get the correct results. For LLM B, the plan ``\textit{find the object first}" can guide it to extract ``\textit{Hotel California}" based on the sentence structure, but if our plan is ``\textit{find the subject first}", the model tends to extract ``\textit{California}" as the location and affect the extraction of object.
Consequently, it is difficult to obtain sufficient information to plan the optimal execution order by relying solely on the LLMs' planning ability.

Therefore, we believe that reinforcement learning is essential to generate plans that are acceptable to complex IE cases.
In this paper, we establish a framework to effectively guide LLMs for multi-step extraction planning, which can have stable and effective extraction performance in different languages, sentence types, and tasks \textbf{without fine-tuning}. We propose a decision model to assist LLMs in executing entity extraction, adaptively deciding the subsequent action(s) based on the current state. We construct the environment with the LLM-based extractor, introduce an indicator as the reward function, and assign reward values based on the semantic correctness and token-level precision of the extraction results.
After training through \textbf{reinforcement learning} and subsequently establishing a Deep Q-Network (DQN), the BERT-based decision model can collaboratively execute the sequential entity extraction tasks with the LLM-based extractor.
In other words, we adopt model collaboration to deal with the entanglement of planning and extracting, let the decision model and the LLM ``\textbf{do what they are good at respectively}" (LLM for language understanding and extracting, and the decision model for fitting Q values and planning), and combine them through RL.

In summary, our main contributions are as follows:

\begin{itemize}
\item First, we design a two-stage multi-step information extraction framework for LLMs, model order selection as Markov Decision Process (MDP), and propose an easy-to-train decision model to guide LLMs in multi-step planning.
\item Second, we adopt RL based framework to train the decision model, which could provide the optimal order for sequential entity extraction. We build the LLM-based extraction environment and design the rewards to reflect the semantic correctness and token-level precision of the results.
\item Third, the experimental results demonstrate the effectiveness of our method. We design an evaluation metric that amalgamates text similarity and exact matching, which is suitable for assessing the extraction capability of LLMs.
\end{itemize}

\section{Related Work}


\subsection{LLM for Information Extraction}
Recent studies on LLMs demonstrate their proficient performance across various downstream tasks, even when provided with only a few examples as prompts \cite{agrawal2022large, jeblick2023chatgpt}. For IE tasks, some studies indicate that with appropriate prompting, LLMs can achieve comparable performance with the supervised methods in zero-shot or few-shot settings of extraction tasks \cite{wei2023zero, gao2023exploring, tang2023does}. \citet{ding2024improving} integrates LLMs with a small evaluation model for RE tasks, addresses the "low recall" issue, and enhances the extraction performance of LLMs.
Furthermore, \citet{chung2022scaling} shows LLMs can produce outstanding performance through supervision and fine-tuning, and \citet{wadhwa-etal-2023-revisiting} suggests that LLMs should be a standard benchmark for some extraction tasks.

\subsection{Reinforcement Learning in IE Tasks}
Previous studies have also employed reinforcement learning (RL) in IE tasks. \citet{narasimhan-etal-2016-improving} enhances information extraction by utilizing RL to integrate external evidence. \citet{takanobu2019hierarchical} implements a hierarchical RL framework to enhance the linkage between entity mentions and relation types. \citet{zeng2018large} uses the relation of entity pairs as distant supervision and guides the training of the relation extractor with RL to address large-scale RE tasks.
The ordered extraction of entities can be modeled as a sequential decision-making problem, so it can be naturally integrated with RL. For example, \citet{zeng2019learning} contemplates the extraction order of relational facts in a sentence and subsequently trains a sequence-to-sequence model using RL, and \citet{huang-etal-2023-adaptive} proposes an RL-based framework that simultaneously trains BERT-based extraction and decision models to generate a suitable extraction order under a given schema.

\section{Methods}
\begin{figure*}[!ht]
\begin{center}
\includegraphics[width=0.9\linewidth]{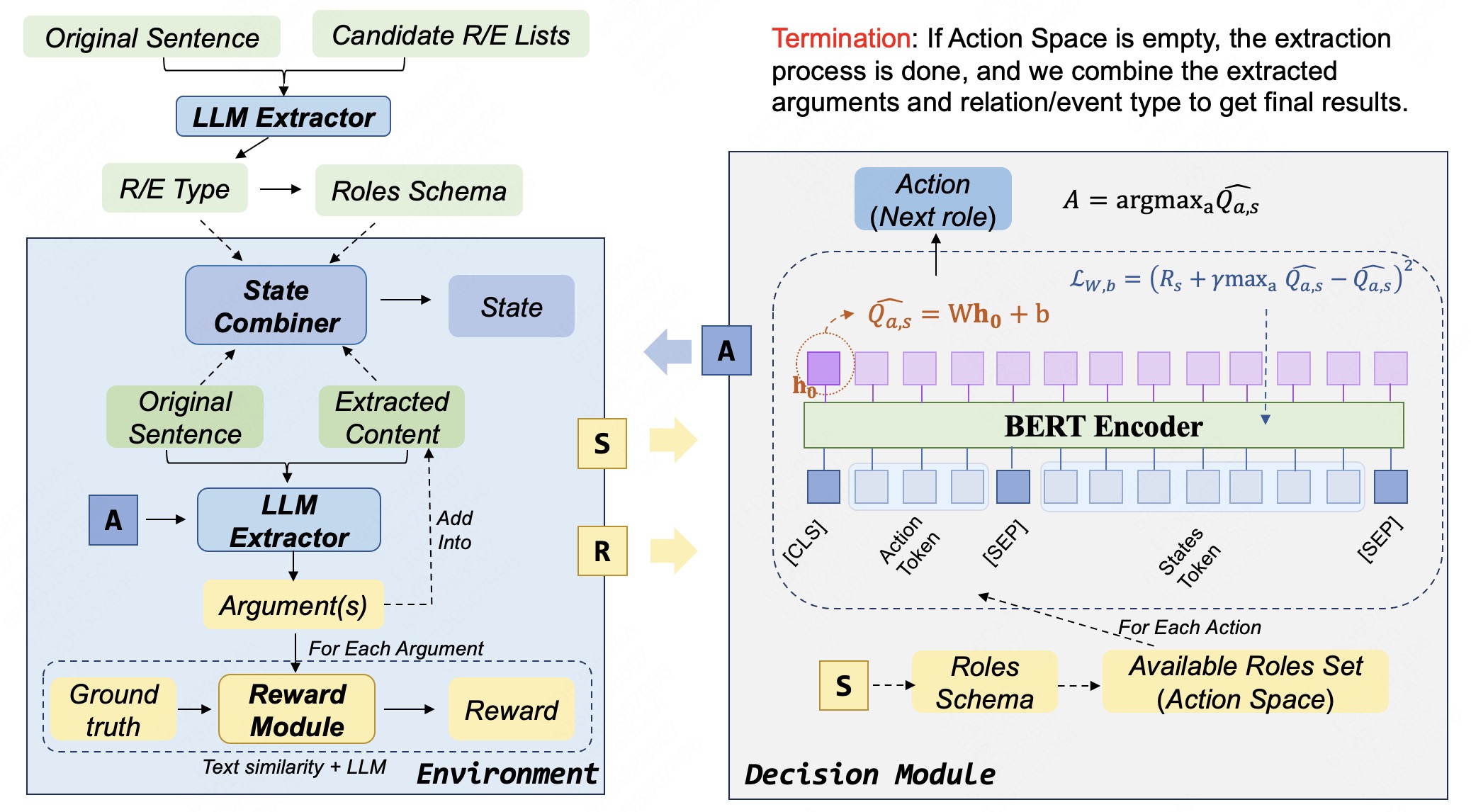} 
\caption{The main workflow of our method. On the left is the extracting process, including relation/event classification and arguments extraction. On the right is the planning part, which guides each step based on the BERT-based Q-Network.}
\label{fig:MDP}
\end{center}
\end{figure*}

\subsection{Preliminary and Task Definition}
Information extraction (IE) tasks are highly diversified due to their different targets \cite{lu-etal-2022-unified}. 
For this paper, we mainly focus on relation extraction (RE) and event extraction (EE) tasks. RE task aims at extracting triples \textit{(subject, predicate, object)} (or \textit{(s,p,o)}) from a given natural language sentence. EE task intends to identify the event types from the given sentence and identify the arguments (entities) corresponding to each role. 
We uniformly define RE and EE as a two-stage task. The goal of the first stage is to classify the relation or event type based on the given sentence, and the second stage is to extract the corresponding entities or arguments. The extracted arguments need to be consistent with the relation/event type and semantically and logically correct.

For the relation/event classification, we provide LLMs with input text and a list of candidate relation/event types and also provide several appropriate examples for different datasets in the prompt to guide LLMs to select the type(s) from the candidate list.
For argument extraction in the second stage, each step only requires LLMs to extract arguments (entities) corresponding to a single role. In the $i$-th step, we provide the input text $s$ and the previously extracted role-argument pairs $(r_j: a_j)_{j=1:i-1}$ as input to LLMs, and instruct LLMs to extract the argument $a_i$ corresponding to the currently required role $r_i$.

\subsection{Solution Framework}
\label{sec:framework}

For different sentences, the order of roles to be extracted should be adaptive rather than fixed. Therefore, we need to consider the previously extracted information instead of directly deciding the extraction order all at once. We train a BERT-based decision model, which determines the role to be extracted next based on the current state (input text and previously extracted information).
We regard the order decision as a \textit{Markov decision process (MDP)} and use reinforcement learning to train the decision model.

In previous work, \citet{huang-etal-2023-adaptive} models BERT-based entity extraction as MDP. However, their decision model and extraction model have the same structure and shared logits, and they are co-trained based on the same training data. Since the fine-tuning of extraction model can also improve the capability, it is not certain whether the final performance improvement comes from the appropriate planning.

In our study, we decouple planning and extracting processes, as shown in Fig.~\ref{fig:MDP}. Generally, we use the LLMs in both stages \textbf{without fine-tuning with labeled data}. For the multi-step argument extraction task, the LLM-based extractor acts as an environment to provide the decision model with the state after the current extraction execution, and the decision model selects the most appropriate role based on the state to guide LLMs to extract step by step until all roles are extracted. The training of the decision model is separated from the execution of the whole extraction process, and the trained decision module can directly interact with similar extraction environments and serve as an instant, universal planner.

The output of LLMs is crucial to completing the RL training of the decision model. To make the output of LLMs more stable, and to enable the LLM-based environment to effectively interact with the decision module, we model the reward assignment as a classification task and design a reward model that takes into account both semantic correctness and token-level matching. The reward model and extraction prompts are described in subsequent subsections.

\subsection{LLM for Relation/Event Classification}
\label{sec:llm_classify}
We directly apply the LLMs as a classifier, which is more straightforward and simpler than training a BERT-based classifier and does not necessitate the provision of additional information such as trigger words. All potential relation/event types within the dataset are compiled into a list and supplied as input, in conjunction with the input text. 

In addition, in some datasets, the relation/event types of the originally labeled data are not natural language expressions. Consequently, it is challenging for LLMs to comprehend the semantics of these relation/event words without undergoing fine-tuning training. To address this, we undertake a natural language conversion for these labels, such as converting "\textit{people-deceased\_person-place\_of\_death}" in NYT10 dataset to "\textit{dead in}". Please refer to Technical Appendix for prompt template.

\subsection{LLM-Based Argument Extractor}
\label{sec:llm_ext}

We adopt an LLM-based extractor to solve the argument extraction task with different extraction orders. For EE tasks, the roles are determined by the event type, and the event type has a corresponding roles schema. The extractor extracts the arguments for the selected role. For RE tasks, the intuitive approach is to set two roles for all relation types: \textit{subject} and \textit{object}. 
However, unlike roles in EE, the words "subject" and "object" do not contain any information about the entity type, which sometimes leads to ambiguity. For example, the subject-object of the relation "affiliated" is person-organization in some datasets, while in other datasets it is department-organization. Therefore, we design the \textbf{roles} in RE tasks as "\textit{subject/object: entity type}". The entity type is determined by the semantics of the selected relation type in the specific dataset.

The input of the arguments extractor is a \textbf{dictionary}, including the original sentence, the current relation/event type, the extracted content (role-argument pairs), and the role (argument type) to extract. At each step of argument extraction, the extractor returns the arguments corresponding to the specified role in a \textbf{string}, separated by commas. The prompt template is provided in Technical Appendix.

\subsection{Environment for Extraction}
We build the environment with the LLM-based extractor, initialize the state based on the relation/event extracted in the first stage, and determine the \textbf{roles schema} (as the initial action space). In each step, the decision model takes the input sentence and extracted content as input and chooses a previously unselected role as the action. The environment receives the action from the decision model, constructs the extractor's input, and performs extraction. After collecting the extraction results, the reward module would assign the reward based on the ground truth of the labeled sample, separate the different arguments of the current role, build the successor state(s), and transit to a new state. When all roles are selected (the action space is empty), the decision model returns a termination signal, and the environment converts the extracted content into structural output. The left of Fig.~\ref{fig:MDP} shows the composition of the environment in detail.

\subsubsection{State}
The state includes the original sentence $s$, relation/event $p$, extracted content $C$, and roles schema $\mathcal{M}$. We denote the state of step $i$ as $S_i$, as follows:
\begin{equation}
	S_i = (\mathcal{M},C_i,p,s)
\end{equation}
Where the extracted content consists of previously extracted role-argument pairs: $C_i = (r_j: a_j)_{j=1:i-1}$. 

\subsubsection{Action}
For a given sentence and corresponding relation/event type, the action of the decision model is the next role to extract. The initial action space $A_0$ is the roles schema $\mathcal{M}$ in the state. In a complete argument extraction process, the extraction order is a permutation of roles in $\mathcal{M}$, and each role should be extracted only once. Therefore, the action space of the decision model is reduced at each step. After selecting action (role) $a_i = r_i$ at the $i$-th step, $a_i$ will be removed from the action space $A_i$ to derive $A_{i+1}$.

\subsection{The Reward Module}
\label{sec:rewards}
The design of rewards is crucial in RL training and is directly related to the loss and the update of model weights.
In our framework, the reward value is used to measure the quality of the extraction results and guide the decision model to make better plans for each episode. 
However, \textbf{the output of a large model is usually uncertain and imprecise}. On one hand, applying the exact match with the ground-truth will misjudge many semantically correct answers as wrong, reducing the effectiveness of the reward value. On the other hand, IE tasks require accurate results in some specific areas and scenarios.
Therefore, we hope to assign high reward values to outputs that are \textbf{both semantically correct and highly matched with the ground-truth}, and use low reward values to guide LLMs to avoid outputs that are semantically incorrect or mismatched with the ground-truth. 
We design an \textit{indicator function} to judge the semantic correctness and apply \textit{text similarity} to remove results with low accuracy.

We set the reward function to an \textbf{indicator function}, that is, to assign a reward value of 1 to acceptable extraction results\footnote{To match the magnitude of the $Q$ value generated by the decision model, in practice we use 0 or 10 as the reward.}, and to assign a reward value of 0 to semantically incorrect or mismatched answers. We directly use an LLM to evaluate the \textbf{semantic correctness}, automatically implement this concise but effective reward function.

Note that our reward function has only two discrete values (which can be regarded as two labels), rather than a continuous function. On one hand, it is intuitive and reasonable to classify the extraction results into "acceptable" and "rejective", but it is difficult to find a reliable and stable method to further sort the acceptable results and assign different rewards. On the other hand, using an LLM as our reward model is convenient and effective. It only needs to judge whether the extraction results are acceptable and output 0 or 1. 

\begin{algorithm}[tb]
\caption{Training of decision model}
\label{algo:main}
\textbf{Input}: Train sets $\mathcal{T}_1,...,\mathcal{T}_T$ with the same task type and language; Initialized replay memory $\mathcal{D}$\\
\textbf{Parameter}: $\theta$: network parameters; $\theta^\prime$: target-net parameters; $N_b$:training batch size; $T_h$: Threshold of text-similarity; hyper-parameters $E,\epsilon,\gamma,k$
\begin{algorithmic}[1] 
\FOR{epoch$=$\rm{1},...,$E$}
\STATE Sample instance $s$ from $T$ labeled train sets, with probability $1/T$ for each set.
\STATE Get the relation or event type $p$, and corresponding roles schema $\mathcal{M}_0$.
\STATE Initialize the state $S_0=(\mathcal{M}_0,C_0,p,s)$.
\FOR{t$=$\rm{1},...,$|\mathcal{M}|$}
\STATE $p =$ \texttt{Random}$(0,1)$
\IF {p $\le$ \rm{1}-$\epsilon$}
\STATE $a_t = \arg\max_a Q(S_t,a;\theta)$
\ELSE
\STATE $a_t =$ \texttt{Random-sample}$(\mathcal{A}_t)$.
\ENDIF
\STATE $aug_{t+1}=$ \texttt{Extractor}$(S_t,a_t)$
\STATE Get ground truth $g_{t+1}$ from instance $s$.
\IF {\texttt{text-similar}$(aug_{t+1},g_{t+1}) \le T_h$}
\STATE $r_{t+1}=0$
\ELSE
\STATE $r_{t+1}=$ \texttt{Reward-Model}$(aug_{t+1},g_{t+1})$
\ENDIF
\STATE Construct $S_{t+1}$, and store $(S_t,a_t,r_{t+1},S_{t+1})$ in $\mathcal{D}$
\STATE Randomly sample $N_b$ transitions $(S_j,a_j,r_j,S_{j+1})$ from $\mathcal{D}$, and do follows:
\IF{$S_{j+1}$ is terminal state}
\STATE $y_j=r_j$
\ELSE
\STATE $y_j=r_j+\gamma \max_{a}Q(S_{j+1},a;\theta^\prime)$
\ENDIF
\STATE $\mathcal{L}(\theta)=(y_j-Q(S_t,a_t;\theta))^2$
\STATE Update parameter $\theta$
\STATE Replace target-net parameters $\theta^\prime$ = $\theta$ every $k$ steps
\ENDFOR
\ENDFOR
\end{algorithmic}
\end{algorithm}

During the RL training process, we use an LLM with a larger scale and stronger capabilities than the extraction model as the reward model to ensure accurate and effective reward assignment. This process also performs \textbf{implicit knowledge distillation} through the rewards. We instruct it to consider whether the semantics of the extracted arguments are consistent with the ground-truth and the token-level accuracy. See Technical Appendix for the prompt template.

We use the \textit{token-level text similarity (Jaccard Similarity)} between ground-truth and extracted argument to measure the token-level accuracy and use it as a threshold to remove "mismatched" extraction results. 
After LLM extraction, we first calculate the similarity score of each argument based on the ground-truth, and set the reward corresponding to the low-score arguments to 0 without passing them to the reward function. This step can be regarded as a pruning process, which reduces the successor state space and enhances the effectiveness when performing inference.
The entire reward module can be formulated as follows:
\begin{equation}
R(x)=
\begin{cases}
1, &\text{if}\ Sim_{x,g} > T\ \text{and}\ Ind_{x,g}=1 \\
0, &\text{otherwise.}
\end{cases}
\end{equation}
Where $g$ is the ground-truth, $Sim_{x,g}$ refers to the text similarity between $x$ and $g$.

\subsection{Decision Module for Planning}
We use a BERT-based decision model to evaluate the value of \textit{(State, Action)} pair to plan the next action for the environment. 
At each step, we combine each role $a_{t_i}$ in the action space with the state $S_t$ as the input sequence: $x_{t_i}$ = [\verb|[CLS]|,$a_{t_i}$,\verb|[SEP]|,$S_t$,\verb|[SEP]|].

The BERT encoder converts these tokens into hidden vectors, and we select the first vector $h_0$ as the representative of the (State, Action) pair. The final evaluation of $Q(S_t,a_t)$ is defined as:
\begin{equation}
	\hat{Q}(S_t,a_t) = \mathbf{W}h_0 + \mathbf{b}
\end{equation}
where $ \mathbf{W}$ and $\mathbf{b}$ are trainable parameters.

We apply \textit{Deep Q-Learning} to train the model and build a Deep Q-Network (DQN) \cite{mnih2013playing}. We design the following learning object:
\begin{equation}
	Q(S,a) = r_{(S,a)} + \gamma \frac{1}{|\mathcal{S}_{(S,a)}|} \cdot \sum_{S^{\prime}\in \mathcal{S}_{(S,a)}}\max_{a^{\prime}\in \mathcal{A}} Q(S^{\prime},a^{\prime})
\end{equation}
where $r_{(S,a)}$ is the reward, $\mathcal{S}_{(S,a)}$ is the set of the successor states derived from the state $S$ with action $a$, and $\gamma$ is the discount factor.

We adopt the DDQN algorithm \cite{van2016deep}, utilize $\epsilon -$greedy exploration and the Experience Replay \cite{mnih2013playing} for RL training. The loss function is defined as the expected value of the mean squared Temporal Difference (TD) error:

\begin{equation}
    \mathcal{L} = \mathbb{E}(\hat{Q}(S,a)-Q(S,a))^2
\end{equation}

Furthermore, for different datasets of the same language and task type (such as different English RE datasets), the state structures of them are the same, so their environments are similar. Therefore, only need to \textbf{mix the training data to train one decision model}, which simplifies the training process of the decision model and improves the generalization of our method. The training process is illustrated in Algorithm \ref{algo:main}.

\section{Experiments}
\label{sec:main_exp}

\begin{table*}[h]
	\centering
	\resizebox{1.4\columnwidth}{!}{
		\begin{tabular}{p{3cm}rrrrrrrrr}
			\toprule
			& \multicolumn{3}{c}{SKE21}  & \multicolumn{3}{c}{DuEE} & \multicolumn{3}{c}{ACE05}\\
			\cmidrule{2-10}    LLM   & \multicolumn{1}{c}{Prec.} & \multicolumn{1}{c}{Reca.} & \multicolumn{1}{c}{F1} & \multicolumn{1}{c}{Prec.} & \multicolumn{1}{c}{Reca.} & \multicolumn{1}{c}{F1} & \multicolumn{1}{c}{Prec.} & \multicolumn{1}{c}{Reca.} & \multicolumn{1}{c}{F1}  \\
			\midrule
		GPT-3.5 & 88.68 & 91.12 & 89.88 & 87.92 & 93.57 & 90.66 & 91.93 & 92.50 & 92.21\\
            Qwen-1.5 & 91.49 & 93.93 & 92.69 & 87.69 & 88.58 & 88.13 & 85.25 & 89.00 & 87.08 \\ 
            Mistral-v0.3 & 79.05 & 88.42 & 83.47 & 81.40 & 87.43 & 84.31 & 88.91 & 90.16 & 89.53 \\
			\bottomrule
		\end{tabular}%
	}
	\caption{The results of relation/event classification of different LLMs on SKE21, DuEE and ACE05.}
	\label{tab:classify_result}%
\end{table*}

\begin{table*}[htb]
	\centering
	\resizebox{1.97\columnwidth}{!}{
		\begin{tabular}{p{0.75cm}p{3cm}rrrrrrrrrrrr}
			\toprule
		&  & \multicolumn{3}{c}{HacRED}  & \multicolumn{3}{c}{SKE21}  & \multicolumn{3}{c}{DuIE}  & \multicolumn{3}{c}{DuEE}     \\
			\cmidrule{3-14}    LLM & Methods & \multicolumn{1}{c}{Prec.} & \multicolumn{1}{c}{Reca.} & \multicolumn{1}{c}{F1} & \multicolumn{1}{c}{Prec.} & \multicolumn{1}{c}{Reca.} & \multicolumn{1}{c}{F1} & \multicolumn{1}{c}{Prec.} & \multicolumn{1}{c}{Reca.} & \multicolumn{1}{c}{F1} & \multicolumn{1}{c}{Prec.} & \multicolumn{1}{c}{Reca.} & \multicolumn{1}{c}{F1} \\
			\midrule
		\multirow{4}{*}{\text{ }\text{ }\rotatebox{90}{GPT-3.5}} & ChatIE & \textbf{81.61} & \textbf{48.51} & \textbf{60.85} & \textbf{71.67} & 43.31 & \textbf{53.99} & \textbf{70.13} & 66.13 & \textbf{68.07} & 57.05 & 78.13 & 65.95 \\
        & LLM + Filter & 22.34 & 25.89 & 23.98 & 43.08 & 51.27 & 46.82 & 58.38 & \textbf{72.56} & 64.70 & - & - & - \\    
        & Ordered -RL (ours) & 32.93 & 32.72 & 32.82 & 45.84 & \textbf{53.64} & 49.43 & 63.53 & 66.99 & 65.21 & \textbf{58.95} & \textbf{79.58} & \textbf{67.73} \\
    & & &  \multicolumn{2}{r}{-46.06\%} &  &  \multicolumn{2}{r}{-8.45\%} &  &  \multicolumn{2}{r}{-4.20\%} &  &  \multicolumn{2}{r}{+2.70\%}  \\
    
    \midrule
		\multirow{6}{*}{\text{ }\text{ }\rotatebox{90}{Qwen-1.5}} & ChatIE & \textbf{55.99} & 19.85 & 29.31 & 48.94 & 35.25 & 40.99 & \textbf{47.40} & 39.14 & 42.87 & 49.19 & 38.06 & 42.91 \\
        & LLM + Filter & 22.34 & 25.89 & 23.98 & 39.86 & 57.11 & 46.95 & 26.42 & 48.08 & 34.10 & - & - & - \\
        & Ordered -random & 31.60 & 34.82 & 33.13 & 67.10 & 70.51 & 68.76 & 31.13 & 46.99 & 37.45 & 53.39 & 60.14 & 56.56 \\
        & Ordered -fixed & 29.82 & 29.07 & 29.44 & 60.68 & 63.31 & 61.97 & 28.47 & 40.75 & 33.52 & 50.38 & 55.83 & 52.96 \\          
        & Ordered -RL (ours) & 36.02 & \textbf{39.98} & \textbf{37.89} & \textbf{76.01} & \textbf{79.75} & \textbf{77.83} & 40.63 & \textbf{57.10} & \textbf{47.47} & \textbf{53.63} & \textbf{62.64} & \textbf{57.78} \\
    & &  & \multicolumn{2}{r}{+14.37\%} &  &  \multicolumn{2}{r}{+13.19\%} &  &  \multicolumn{2}{r}{+9.10\%} &  &  \multicolumn{2}{r}{+2.11\%}  \\

   \midrule
        \multirow{4}{*}{\text{ }\text{ }\rotatebox{90}{\small Mistral-v0.3}} & ChatIE & \textbf{34.56} & 7.09 & \textbf{11.76} & \textbf{31.50} & 8.73 & 13.67 & \textbf{42.84} & 31.18 & 36.09 & 37.73 & 11.94 & 18.14 \\
        & LLM + Filter & 6.03 & \textbf{14.37} & 8.49 & 28.81 & \textbf{38.60} & \textbf{32.99} & 36.53 & \textbf{37.27} & 36.90 & - & - & - \\    
        & Ordered -RL (ours) & 10.40 & 12.07 & 11.17 & 27.04 & 32.29 & 29.43 & 40.05 & 36.34 & \textbf{38.10} & \textbf{54.21} & \textbf{69.72} & \textbf{61.00} \\
    & & &  \multicolumn{2}{r}{-5.02\%} &  &  \multicolumn{2}{r}{-10.79\%} &  &  \multicolumn{2}{r}{+3.25\%} &  & \multicolumn{2}{r}{+236.27\%}  \\
		\bottomrule
		\end{tabular}%
	}
	\caption{The main evaluation results of different methods on Chinese IE datasets. Better precision, recall and F1 score are marked \textbf{bold}. We also report the percentage improvement on F1 score.}
	\label{tab:mainresult_chinese}%
\end{table*}%

\subsection{Datasets}
We evaluate our methods on several public and accessible complicated information extraction datasets, including the relation extraction datasets NYT~\cite{riedel2010modeling}, NYT11-HRL~\cite{takanobu2019hierarchical}, Wiki80~\cite{han2019opennre}, SKE21~\cite{xie-etal-2021-revisiting}, HacRED~\cite{hacred} and DuIE~\cite{li2019duie}, and the event extraction dataset DuEE~\cite{li2020duee} and ACE05\footnote{\url{https://catalog.ldc.upenn.edu/LDC2006T06}}. These datasets are challenging for LLMs and widely used for evaluation of existing extraction methods. A brief introduction to these datasets is provided in Technical Appendix.

\subsection{Comparing Methods and Metrics}
We select some recently popular and high-performance LLMs as extractors in our experiments. We choose the open-source model Mistral-7B-instruct-v0.3\footnote{\url{https://huggingface.co/mistralai/Mistral-7B-Instruct-v0.3}}, Qwen1.5-14B~\cite{bai2023qwen}, and the famous GPT3.5-turbo \cite{brown2020language, ouyang2022training} for both English and Chinese datasets. We apply two previous multi-stage extraction methods to our testset, ChatIE~\cite{wei2023zero} and LLM(without fine-tuning)+Filter~\cite{ding2024improving} and compare our method with them. We also compare our method with random/fixed selection of actions, and specifically calculate the metrics in cases involving multiple triples or roles.

We design a relaxed evaluation method based on \textbf{the exact match}. We consider the correctness of the arguments at the token level and calculate the similarity between the extraction result and ground truth for each argument (according to Sec.~\ref{sec:rewards}). In the level of the entire triple or event, we believe the extraction result \textbf{matches} the ground truth only if the relation/event type is classified correctly and the text similarity is greater than the threshold for all arguments. We calculate and report precision (Prec.), recall (Reca.), and F1 scores for all the experiments according to the method above. We repeat the experiment 3 times and averaged the metrics as final results.
See Technical Appendix for details.

\subsection{Effectiveness of LLM Classification}
We first assess the efficacy of LLM-based relation/event classification on the SKE21, DuEE and ACE05 datasets. Table~\ref{tab:classify_result} shows that LLMs classify the relation/event accurately and achieve stable and accurate results, setting the stage for the argument extraction.

\begin{table*}[htb]
	\centering
	\resizebox{1.97\columnwidth}{!}{
		\begin{tabular}{p{0.75cm}p{3cm}rrrrrrrrrrrr}
			\toprule
			& & \multicolumn{3}{c}{NYT10}  & \multicolumn{3}{c}{NYT11-HRL}  & \multicolumn{3}{c}{Wiki80}  & \multicolumn{3}{c}{ACE05}     \\
			\cmidrule{3-14}      LLM & Methods & \multicolumn{1}{c}{Prec.} & \multicolumn{1}{c}{Reca.} & \multicolumn{1}{c}{F1} & \multicolumn{1}{c}{Prec.} & \multicolumn{1}{c}{Reca.} & \multicolumn{1}{c}{F1} & \multicolumn{1}{c}{Prec.} & \multicolumn{1}{c}{Reca.} & \multicolumn{1}{c}{F1} & \multicolumn{1}{c}{Prec.} & \multicolumn{1}{c}{Reca.} & \multicolumn{1}{c}{F1} \\
			\midrule
			\multirow{4}{*}{\text{ }\text{ }\rotatebox{90}{GPT-3.5}} & ChatIE & 46.27 & 42.70 & 44.41 & 32.12 & 42.97 & 36.76 & \textbf{32.14} & 29.57 & 30.80 & \textbf{57.52} & 22.18 & 32.02 \\
             & LLM + Filter & 39.12 & 44.06 & 41.44 & \textbf{33.06} & 47.19 & 38.88 & 17.08 & 18.15 & 17.60 & - & - & - \\    
            & Ordered -RL (ours) & \textbf{52.31} & \textbf{68.84} & \textbf{59.44} & 32.98 & \textbf{49.19} & \textbf{39.49} & 31.90 & \textbf{48.29} & \textbf{38.42} & 25.57 & \textbf{65.19} & \textbf{36.73} \\
    & &  & \multicolumn{2}{r}{+33.84\%} &  &  \multicolumn{2}{r}{+1.57\%} &  &  \multicolumn{2}{r}{+21.94\%} &  &   \multicolumn{2}{r}{+14.71\%}  \\
    
    \midrule
		\multirow{4}{*}{\text{ }\text{ }\rotatebox{90}{Qwen-1.5}} & ChatIE & 39.66 & 15.66 & 22.45 & 27.48 & 19.46 & 22.78 & 16.67 & 18.29 & 17.44 & 41.38 & 12.29 & 18.95 \\
        & LLM + Filter & 12.36 & 19.22 & 15.04 & 25.82 & 29.97 & 27.74 & 6.64 & 9.50 & 7.82 & - & - & - \\    
        & Ordered -RL (ours) & \textbf{62.62} & \textbf{45.54} & \textbf{52.73} & \textbf{30.68} & \textbf{35.41} & \textbf{32.87} & \textbf{31.18} & \textbf{36.57} & \textbf{33.66} & \textbf{39.58} & \textbf{77.82} & \textbf{52.47} \\
   & &  &  \multicolumn{2}{r}{+134.88\%} &  &  \multicolumn{2}{r}{+18.49\%} &  &  \multicolumn{2}{r}{+93.00\%} &  &  \multicolumn{2}{r}{+176.89\%}  \\

   \midrule
		\multirow{6}{*}{\text{ }\text{ }\rotatebox{90}{Mistral-v0.3}} & ChatIE & 25.61 & 11.84 & 16.19 & 19.38 & 13.51 & 15.92 & 6.18 & 9.14 & 7.37 & \textbf{51.16} & 47.51 & 49.27 \\
        & LLM + Filter & 38.50 & 59.91 & 46.88 & 16.67 & 19.41 & 17.94 & 8.34 & 9.92 & 9.06 & - & - & - \\  
        & Ordered -random & 49.73 & 83.22 & 62.26 & 23.57 & 54.18 & 32.85 & 18.18 & 35.43 & 24.03 & 28.03 & 81.23 & 41.68 \\
        & Ordered -fixed & 47.87 & 80.82 & 60.13 & 23.26 & \textbf{59.73} & 33.48 & 16.01 & 32.00 & 21.34 & 30.39 & 89.42 & 45.37 \\
         & Ordered -RL (ours) & \textbf{50.47} & \textbf{85.02} & \textbf{63.34} & \textbf{23.70} & 59.19 & \textbf{33.85} & \textbf{20.43} & \textbf{40.29} & \textbf{27.12} & 39.94 & \textbf{94.88} & \textbf{56.22} \\
    & &  & \multicolumn{2}{r}{+1.73\%} &  &  \multicolumn{2}{r}{+1.11\%} &  &  \multicolumn{2}{r}{+12.86\%} &  &  \multicolumn{2}{r}{+14.11\%}  \\
			\bottomrule
		\end{tabular}%
	}
	\caption{The main evaluation results of different methods on English IE datasets. Better precision, recall and F1 score are marked \textbf{bold}. We also report the percentage improvement on F1 score.}
	\label{tab:mainresult_english}%
\end{table*}%

\subsection{Main Results}

Table~\ref{tab:mainresult_chinese} and ~\ref{tab:mainresult_english} show the extraction results on the Chinese and English datasets respectively. 
ChatIE~\cite{wei2023zero} uses a multi-step instruction to guide ChatGPT to extract, we also apply this instruction template to the open source model and report the results, as shown in the first row of each block. We find that the original paper uses zero-shot to conduct experiments on ChatGPT, but for open-source LLMs other than ChatGPT, such zero-shot prompts produced poor results. Therefore, we used the multi-stage prompt template of ChatIE, but added the same number of in-context examples as our method to conduct baseline experiments.
\citet{ding2024improving} adds the filtered results of small models to the prompt of LLMs\footnote{\citet{ding2024improving} use the large and small models to collaboratively perform complex triple extraction tasks. They trained a small model that can filter out "truly related entity pairs" and prompted the filtered results to the large model.}, and we report the performance of this method on our test set, as shown in the second row of each block. None of the LLMs in the table are fine-tuned on a labeled training set.

The results indicate that, in most cases, our method has better extraction capabilities than the other two prompt-based fixed-ordered planning methods. 
Previous methods usually only get good results on specific models or datasets, and lack generalization ability. For example, the results of ChatIE on open-source models are far worse than the results on GPT3.5-turbo, and in most cases when ChatIE achieves the best results, our method also gets high recall and F1 score.

In general, for various types of LLM extractors and IE datasets, our method can achieve the highest precision, recall, and F1 score in most cases. Our RL-based multi-step planning method is effective and generalizable.

\begin{table}[h]
	\centering
	\resizebox{0.98\columnwidth}{!}{
		\begin{tabular}{p{0.75cm}p{3cm}rrrrrr}
			\toprule
		& & \multicolumn{3}{c}{SKE21}  & \multicolumn{3}{c}{DuEE} \\
			\cmidrule{3-8}    LLM & Methods  & \multicolumn{1}{c}{Prec.} & \multicolumn{1}{c}{Reca.} & \multicolumn{1}{c}{F1} & \multicolumn{1}{c}{Prec.} & \multicolumn{1}{c}{Reca.} & \multicolumn{1}{c}{F1}  \\
			\midrule
			\multirow{5}{*}{\text{ }\text{ }\rotatebox{90}{GPT-3.5}} & ChatIE & 17.37 & 28.56 & 21.60 & 42.31 & 65.38 & 51.37  \\
            & Ordered -random & 19.07 & 33.87 & 24.40 & 55.48 & 80.57 & 65.71  \\   
            & Ordered -fixed & 6.62 & 11.75 & 8.47 & 48.16 & 67.14 & 56.09  \\
            & Ordered -RL (ours) & \textbf{24.58} & \textbf{40.09} & \textbf{30.47} & \textbf{59.39} & \textbf{86.62} & \textbf{70.47}  \\
    &  &  & \multicolumn{2}{r}{+24.88\%} &  &  \multicolumn{2}{r}{+7.24\%}  \\

   \midrule
			\multirow{5}{*}{\text{ }\text{ }\rotatebox{90}{Qwen-1.5}} & ChatIE & 25.31 & 14.06 & 18.07 & 48.72 & 23.75 & 31.93  \\
            & Ordered -random & 55.44 & 63.36 & 59.14 & 60.23 & 67.52 & 63.66  \\
            & Ordered -fixed & 49.79 & 55.76 & 52.61 & 57.27 & 61.46 & 59.29  \\
            & Ordered -RL (ours) & \textbf{63.69} & \textbf{72.35} & \textbf{67.75} & \textbf{65.19} & \textbf{79.94} & \textbf{71.82}  \\
    &  &  & \multicolumn{2}{r}{+14.56\%} &  &  \multicolumn{2}{r}{+11.36\%}  \\
			\bottomrule
		\end{tabular}%
	}
	\caption{The extraction results of complicated situations on SKE21 and DuEE. We select sentences with more than 4 triples from the SKE21, and events with more than 5 roles from the DuEE.}
	\label{tab:result_complex}%
\end{table}%

\subsection{Results of Different Step Orders}
To verify the effectiveness of our decision module, we also compared the extraction results of different order selection methods for entity extraction, using Qwen for Chinese data and Mistral for English data, as shown in Table~\ref{tab:mainresult_chinese} and Table~\ref{tab:mainresult_english}. The term "-random" means randomly selecting actions when extracting arguments, and "-fixed" means using the same order for all actions (e.g., \textit{subject} first, then \textit{object}).
In most cases, RL-based extraction order can achieve the best results, and the metrics of fixed extraction order are lower than random order. These results imply that the optimal order varies across distinct instances, and our method could adaptively make the best plan.

\subsection{Complicated Extraction Settings}
To further demonstrate the stability and effectiveness of our method and reflect the advantages of dynamic RL-based order selection, we choose more complex situations for experiments, which contain multiple triples or roles. We select sentences with more than 4 triples from the SKE21 test set, and events with more than 5 roles from the DuEE test set to conduct experiments.

The results of Table~\ref{tab:result_complex} show that when the IE task becomes complicated, the extraction performance of ChatIE is far inferior to our method. In addition, compared with random or fixed extraction order, our method achieves significant improvements in the metrics. 
This also suggests that when LLMs undertake extraction tasks in complex scenarios, an appropriate order can further augment the model's extraction ability.

\section{Conclusion}
In this paper, we propose a two-stage multi-step method for information extraction tasks of LLMs. We use LLM extractors as the environment, design a reward module that takes into account both semantic correctness and token-level matching, and employ a reinforcement learning framework to train the decision model, which can adaptively select the optimal extraction order.

Experimental results show that our method performs better than fixed-ordered planning in most cases and can be applied to various LLMs, consistently enhancing their IE capabilities. By comparing different order selection methods, it is also verified that our RL-based framework can adaptively perform multi-step extraction planning, bringing stable effect improvement in both general and complex situations. 

\newpage
\bibliography{aaai25}

\newpage

\end{document}


\maketitle

%
\appendix

\section{Dataset Introduction}
\label{sec:dataset}

\paragraph*{NYT series}
NYT is based on the articles in New York Times. There are many derived datasets with better labeling. NYT10~\cite{riedel2010modeling} and NYT11~\cite{Hoffmann2011KnowledgeBasedWS} label the complete entities. Moreover, NYT10-HRL and NYT11-HRL~\cite{takanobu2019hierarchical} are better versions that are processed by optimizing the relation labels.

\paragraph*{Wiki80}
Wiki80~\cite{han2019opennre} is a sentence-level dataset based on the few-shot dataset FewRel~\cite{han2018fewrel}, which contains 80 types of relations. The number of each relation is 700, with a total of 56,000 samples.

\paragraph*{HacRED} HacRED~\cite{hacred}\footnote{\url{https://github.com/qiaojiim/hacred}} is a novel challenging extraction dataset. It analyzes the performance gap between popular datasets and practical applications, and carefully selects and designs more hard cases. HacRED consists of 65,225 relational facts annotated from 9,231 wiki documents with sufficient and diverse hard cases, which poses a very high challenge to many current complex extraction methods.

\paragraph*{SKE21}
SKE19\footnote{\url{http://ai.baidu.com/broad/download?dataset=sked}} is published by Baidu, and is currently the largest dataset available for complex relational triple extraction. Since its testing set is unpublished, and there are some errors in the validation set, a version named SKE21 is published by \citet{xie-etal-2021-revisiting}. The testing set of SKE21 is carefully manually relabeled and contains 1,150 sentences and 2,765 annotated triples.

\paragraph*{DuIE}
DuIE \cite{li2019duie} is a Chinese information extraction dataset. It contains more than 210, 000 sentences and 48 pre-defined schema gathered from Baidu Encyclopedia, Baidu Tieba and Baidu Information Stream. Compared to the previous version, it contains 5 Multiple-O-values schema (Schema with multiple object slots and values), which greatly increase the difficulty of the task.

\paragraph*{DuEE}
DuEE \cite{li2020duee} is a Chinese document-level event extraction dataset. It contains 11, 224 documents categorized into 65 event types, along with 41, 520 event arguments mapped to 121 argument roles, which is the largest Chinese EE dataset.

\paragraph*{ACE05}
ACE 2005\footnote{\url{https://catalog.ldc.upenn.edu/LDC2006T06}} Multilingual Training Corpus was developed by the Linguistic Data Consortium (LDC) and contains approximately 1,800 files of mixed genre text in English, Arabic, and Chinese annotated for entities, relations, and events. In this paper we use the English events corpus, it provides event annotations in document and sentence levels from a variety of domains such as newswires and online forums.

\section{The Inputs of LLMs}
\label{sec:inputs}
In the methods section of the main submission, we introduce the inputs for the LLM-based classifier and extractor respectively. We describe the natural language conversion of relation/event types, and the design of roles in RE tasks when extracting arguments. Here we provide some illustrative intuitive examples based on the NYT11 dataset.
\subsection{Natural Language Conversion of Types}
We convert the original relation words in the dataset into natural language expressions that are conducive to the comprehension of LLMs. For example,
\begin{itemize}
    \item[*] /location/location/contains - location contains
    \item[*] /business/person/company - work at
    \item[*] /people/person/place-of-birth - born in
    \item[*] /business/company/founders - founder
\end{itemize}

\subsection{Roles in RE Tasks}
We design the roles in RE tasks as "\textit{subject/object: entity type}" to reduce the ambiguity of relation types and promote the argument extraction of LLMs. For example, in NYT11 datasets:
\begin{itemize}
    \item[*] the roles of relation "location contains" are: "\textit{subject: place}" and "\textit{object: place}"
    \item[*] the roles of relation "work at" are: "\textit{subject: person}" and "\textit{object: company}"
    \item[*] the roles of relation "born in" are: "\textit{subject: person}" and "\textit{object: place}"
    \item[*] the roles of relation "founder" are: "\textit{subject: company}" and "\textit{object: person}"
\end{itemize}

\section{Prompt Templates}
\label{sec:prompts}

\noindent\textbf{Prompt template for classification}: \\
\noindent\texttt{Given the origin sentence, and a relation/event type list $L$, Please select the relation/event contained in this sentence from the list $L$ . If more than one relation/event in $L$ is included in this sentence, please separate them with commas(',') and output.} \\
\noindent\texttt{Here are some examples:} \\
\noindent\texttt{Input: } $s_1,L_1$ \\
\noindent\texttt{Output: } $o_1$ \\
\noindent\texttt{...} \\
\noindent\texttt{Now given the following sentence and relation/event list, please select all relations or events contained in the sentence, and output them.} \\
\noindent\texttt{Note: Consider only the types provided in the list.} \\

\noindent\textbf{Prompt template for entity extraction}: \\
\noindent\texttt{Given the text $s$ and the interested relation/event $p$, we may also give the extracted content $C$. Please identify the argument of the role $r$ according to the given content.} \\
\noindent\texttt{Here are some examples:} \\
\noindent\texttt{Input: } \{$s_1,p_1,C_1$, \texttt{"the role I want"}: $r_1$\} \\
\noindent\texttt{Output: } $a_1$ \\
\noindent\texttt{...} \\
\noindent\texttt{Now given the following input, please complete the extracting task.} \\
\noindent\texttt{Note: The output content should only contains the arguments of the role I want, and needs to match the relation/event type and extracted content provided in the input. If there are multiple corresponding arguments, please separate them with commas(',') and output.} \\

\noindent\textbf{Prompt template for reward assignment}: \\
\noindent\texttt{Now there is a model extraction competition. Please judge whether the answer extracted by the model is correct based on the ground-truth and give a score. If the extraction is correct, please give a score of 1, otherwise 0. } \\
\noindent\texttt{Here are some examples:} \\
\noindent\texttt{Extracted: USA; Ground-truth: the United States; Output: 1} \\
\noindent\texttt{Extracted: Beijing City; Ground-truth: Nanjing City; Output: 0} \\
\noindent\texttt{...} \\
\noindent\texttt{You only need to output the score value 1 or 0, without any redundant output.} \\
\noindent\texttt{Note: The extraction result does not have to be exactly the same as the ground-truth. It can be judged as correct if the meaning is consistent with the ground-truth and does not cause confusion.} \\
\noindent\texttt{Please give a score to the following extracted answer.}

\section{Experiment Details}
\label{sec:experiments}

Our experiments are conducted on three A800 GPUs. All deep models, including the extraction model and decision model, are implemented using the PyTorch framework. For all parts involving the LLMs, we provide 3 examples in the prompt, directly using LLMs to extract without fine-tuning with labeled data. The time out for all LLMs is set to 6s. As for the RL training of the decision module, we set the buffer size to 5000 and the target network update step to 20. For the reward model, we choose Qwen1.5-72B for both Chinese and English datasets.

\begin{table}[!htb]
\small
\centering
	\begin{tabular}{ccc}
		\toprule
		\textbf{Hyper-parameters} & LLMs & Decision Model \\ 
		\midrule 
		Batch size & - & 32   \\ 
		Max length & 2048 & 512  \\ 
		Learning rate & - & 1e-4 \\ 
		Temperature & 0 & -  \\ 
        Top p & 1 & -  \\
		\bottomrule
	\end{tabular}
	\caption{Hyper-parameters for LLMs and decision model.}
	\label{tab:parameters}
\end{table}

To reduce time consumption, for each dataset, we randomly select 2000 items from the training set for the training of the decision model. We train 10 epochs for all datasets. Additionally, the exploration parameter $\epsilon$ is initialized at 0.9 and the discount factor $\gamma$ is set to 0.5. The exploration rate $\epsilon$ becomes 0.9 times its own in every 100 steps until it reaches 0.05.
We employed AdamW optimizer as the optimizer,
with using linear scheduling with warming up proportion 10\%. 

For testing, we randomly select 800 items from the test set for each dataset, and the test data is the same for all methods and LLMs.
The text similarity threshold for "matched (positive) results" mentioned in the experiment section is set to 0.85. In fact, our case analysis find that most of the results judged to be correct are exactly matched with the ground truth.

More details are listed in Table~\ref{tab:parameters}.

\section{Complexity of our method}
\label{sec:complexity}

As mentioned in the methods section, for the same language and type of tasks, we only need to train one decision model. This is efficient and after training, such a model also shows good generalization ability. After the RL training is completed, the inference time of multi-step planning is close to that of the prompt-based method. Indeed, LLMs are more time-consuming than small models, and in simple IE tasks, using a fine-tuned small model is sufficient to obtain good results. However, in complex situations, the performance of LLMs' extraction based on planning is significantly better than that of a small model. Therefore, we suggest that in practical applications, users can choose a suitable extraction method based on the complexity of the task.

\newpage
\bibliography{aaai25}